\documentclass{article}
\usepackage{spconf,amsmath,graphicx}

\usepackage{hyperref}
\usepackage{float}
\usepackage{algorithmic}
\usepackage{textcomp}
\usepackage{amssymb,amsfonts}
\usepackage[numbers,sort&compress]{natbib}

\title{SAM-guided Enhanced Fine-grained Encoding with Mixed Semantic Learning for Medical Image Captioning}

\names{Zhenyu Zhang$^*$\textsuperscript{1}, 
Benlu Wang$^*$\textsuperscript{1},
Weijie Liang$^*$\textsuperscript{1},
Yizhi Li$^*$\textsuperscript{1},
Xuechen Guo\textsuperscript{1},
Guanhong Wang\textsuperscript{1},}
\name{
Shiyan Li\textsuperscript{2},
Gaoang Wang\textsuperscript{1}
\thanks{$^*$ Equal contribution.}}
\address{\textsuperscript{1}Zhejiang University-University of Illinois Urbana-Champaign Institute, Zhejiang University, China\\
\textsuperscript{2}Sir Run Run Shaw Hospital, Zhejiang University School of Medicine, China}

\begin{document}
%
\maketitle
\begin{abstract}
With the development of multimodality and large language models, the deep learning-based technique for medical image captioning holds the potential to offer valuable diagnostic recommendations. However, current generic text and image pre-trained models do not yield satisfactory results when it comes to describing intricate details within medical images. In this paper, we present a novel medical image captioning method guided by the segment anything model (SAM) to enable enhanced encoding with both general and detailed feature extraction. In addition, our approach employs a distinctive pre-training strategy with mixed semantic learning to simultaneously capture both the overall information and finer details within medical images. We demonstrate the effectiveness of this approach, as it outperforms the pre-trained BLIP2 model on various evaluation metrics for generating descriptions of medical images.
\end{abstract}
\vspace{-0.2cm}
\section{Introduction}
\label{sec:intro}
\vspace{-0.1cm}
Medical imaging, including techniques such as ultrasound, MRI, and CT, involves the use of radiographic methods for the diagnosis and treatment of internal human ailments. 
Utilizing deep learning for medical image description technology enables the generation of medical reports, unveiling rich clinical information to offer doctors valuable diagnostic assistance \cite{wang2022medclip, zhang2022contrastive,wu2023medklip}.
In recent years, with the exploration and development of deep learning techniques in the field of text-image multimodality, there has emerged a trend of using contrastive learning for cross-modal pre-training with image and text pairs \cite{li2021align,radford2021learning,li2022blip,li2023blip2}. These models have achieved promising results in downstream tasks such as image captioning. 

Overlooking the detailed semantic information is a main challenge when adapting the general pre-trained models to the medical domain.
In the case of medical imaging, a significant portion of diagnoses relies on the detailed features within specific regions of an image. Due to the inherent ambiguity of language combined with varying levels of granularity in textual descriptions, text-only supervised general models struggle to efficiently capture the fine details present in the images. This limitation arises from the difficulty of the image encoder to capture subtle feature differences in localized areas of images with blurry boundaries, noise, and poor contrast \cite{wang2022medclip}. 
In an effort to enhance model performance, the challenge of unstable generalization capability persists \cite{survey}.

\begin{figure}[!t]
    \centering
    \includegraphics[width=0.5\textwidth]{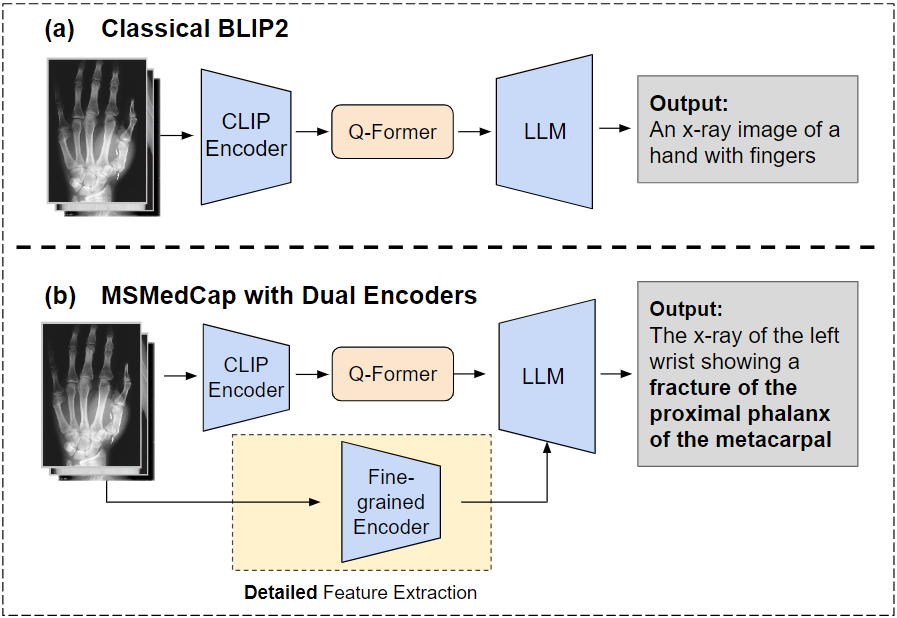}
    \caption{\textbf{Overview of (a) BLIP2 and (b) MSMedCap (Ours).} 
MSMedCap implements a supplemented fine-grained feature extraction to synergize functionalities between encoders.}
\label{fig:overview}
\vspace{-0.4cm}
\end{figure}

In this paper, we propose a \textbf{\underline{Med}}ical image \textbf{\underline{Cap}}tioning model based on a dual-image encoder and \textbf{\underline{M}}ixed \textbf{\underline{S}}emantic learning, namely \textbf{MSMedCap}. 
As shown in \textbf{Fig. \ref{fig:overview}}, MSMedCap employs a detailed feature extraction to capture the fine-grained information.
Specifically, our model contains a dual-encoder architecture: a ViT image encoder pre-trained with CLIP \cite{radford2021learning} to extract the overall information, and a segment anything model (SAM) \cite{kirillov2023segment} guided encoder to capture fine-grained details. 
The functionalities of the two encoders are well complemented and synergized by employing a distinctive pre-training strategy with mixed semantic learning to simultaneously capture both the overall information and finer details within medical images.
We conducted experiments on various datasets \cite{lin2015microsoft,subramanian2020medicat,pelka2018radiology} to evaluate our model, confirming the effectiveness of our proposed method. 

\begin{figure*}[!t]
    \centering
    \includegraphics[width=0.7\textwidth]{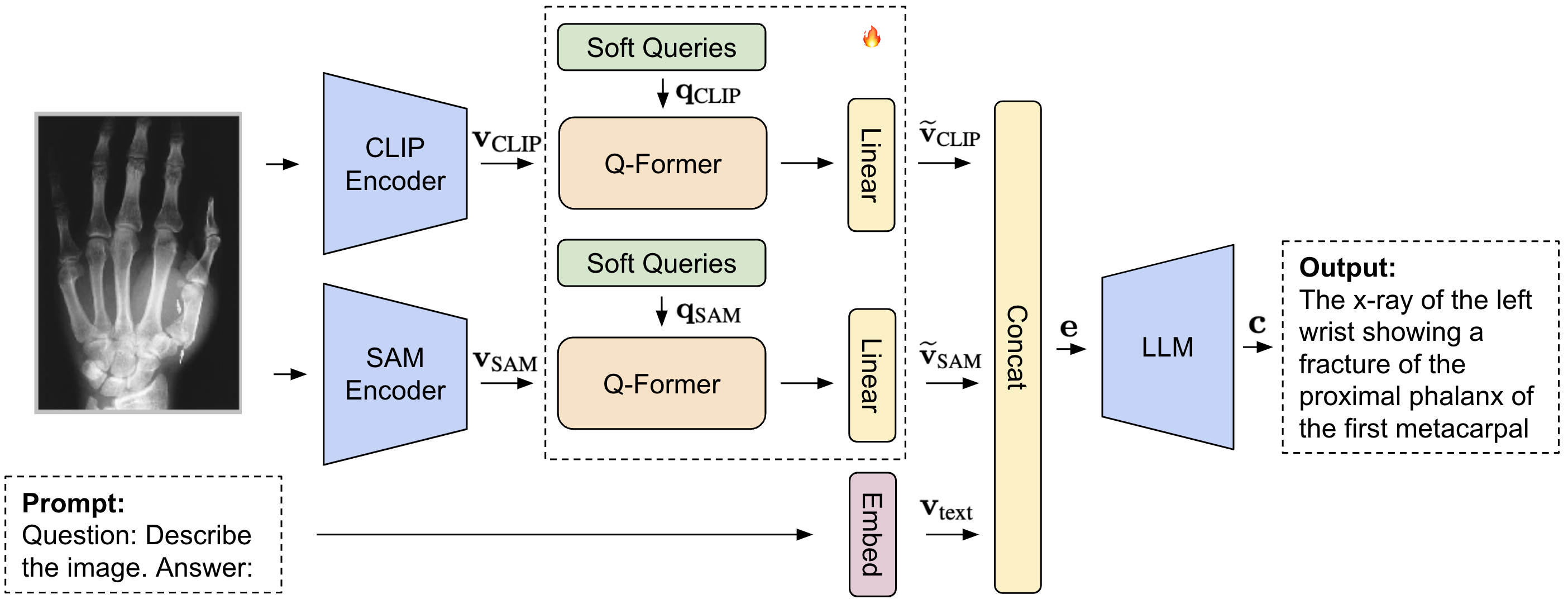}
    \caption{\textbf{Model Architecture.} Utilizing frozen dual image encoders and trainable dual Q-Formers enables the extraction of image features at different granularities. During the full model training, only the parameters of the Q-Formers, Soft Queries, and Linear Layers can be trained.}
    \label{fig:model}
    \vspace{-0.2cm}
\end{figure*}


The key advantages of MSMedCap include:

\vspace{-0.2cm}
\begin{itemize}
\setlength{\itemsep}{0pt}
\setlength{\parsep}{0pt}
\setlength{\parskip}{1pt}
\item We present an innovative SAM-guided dual-encoder architecture that enables the capture of information with different granularities in medical images.
\item A distinctive pre-training strategy is designed to complement both encoders, showcasing excellent synergy while simultaneously preserving the overall and detailed semantic knowledge from the two encoders.
\item Our proposed MSMedCap outperforms the baseline BLIP2 models \cite{li2023blip2} on ROCO \cite{pelka2018radiology} and MedICaT \cite{subramanian2020medicat} datasets with significant improvements.
\end{itemize}

\section{Method}
\label{sec:method}

In this section, we first describe the model architecture. Then we present the mixed semantic pre-training strategy, followed by the illustration of adapting the large language model (LLM) for medical image caption in the down-stream task.

\subsection{Model Architecture}


To capture both general and detailed features, we use the dual-encoder for feature extraction, guided by the segment anything model (SAM) \cite{kirillov2023segment}. Then the features are aligned and aggregated by the dual query Transformers (Q-Former) \cite{li2023blip2} and linear projection layer. Finally, the LLM is adopted to generate medical captions with the text prompt. The overview of the model architecture is shown in \textbf{Fig. \ref{fig:model}}. The details of each component are described as follows.


\noindent \textbf{Dual Image Encoders.}  
We utilize two ViT Encoders trained separately based on CLIP and SAM, namely, $f_{\text{CLIP}}$ and $f_{\text{SAM}}$, to encode image features. The images $\mathbf{x}$ are separately inputted into these two Image Encoders, resulting in two sets of distinct image embedding vectors $\mathbf{v}_{\text{CLIP}} \in \mathbb{R}^{N \times C},\mathbf{v}_{\text{SAM}} \in \mathbb{R}^{Q \times S}$,
\begin{equation}
    \mathbf{v}_{\text{CLIP}} = f_{\text{CLIP}}(\mathbf{x}),
\end{equation}
\begin{equation}
    \mathbf{v}_{\text{SAM}} = f_{\text{SAM}}(\mathbf{x}),
\end{equation}
where $N$ and $Q$ represent the number of feature vectors, $C$ and $S$ represent the dimensions of each feature vector.

\noindent
\textbf{Dual Query Transformers (Q-Former).}  The features outputted by the dual encoders are processed through cross attention separately, resulting in aligned features through their respective Q-Formers ($g_{\text{CLIP}}$ and $g_{\text{SAM}}$) as follows,
\begin{equation}
    \widetilde{\mathbf{v}}_{\text{CLIP}} = g_{\text{CLIP}}(\mathbf{q}_{\text{CLIP}},\mathbf{v}_{\text{CLIP}}),
\end{equation}
\begin{equation}
    \widetilde{\mathbf{v}}_{\text{SAM}} = g_{\text{SAM}}(\mathbf{q}_{\text{SAM}},\mathbf{v}_{\text{SAM}}),
\end{equation}
where $\mathbf{q}_{\text{CLIP}}, \mathbf{q}_{\text{SAM}} \in \mathbb{R}^{M \times D}$ are two sets of learnable query vectors within the Q-Former. Note that linear projection layers are adopted as the output layer of Q-Formers. 



\noindent
\textbf{Large Language Model (LLM).} The OPT LLM \cite{zhang2022opt} is employed to generate medical captions. We combine the outputs of Q-Formers and text prompt embedding $\mathbf{v}_{\text{text}}$ as the input to the LLM, which can be formulated as follows, 
\begin{equation}
    \mathbf{e} = \text{Concat}(\widetilde{\mathbf{v}}_{\text{CLIP}},\widetilde{\mathbf{v}}_{\text{SAM}},\mathbf{v}_{\text{text}}), 
\end{equation}
\begin{equation}
    \mathbf{c} = \text{LLM}(\mathbf{e}),
\end{equation}
where $\text{Concat}()$ represents the concatenation operation.

\subsection{Mixed Semantic Pre-training}

\begin{figure}[!t]
    \centering
    \includegraphics[width=0.35\textwidth]{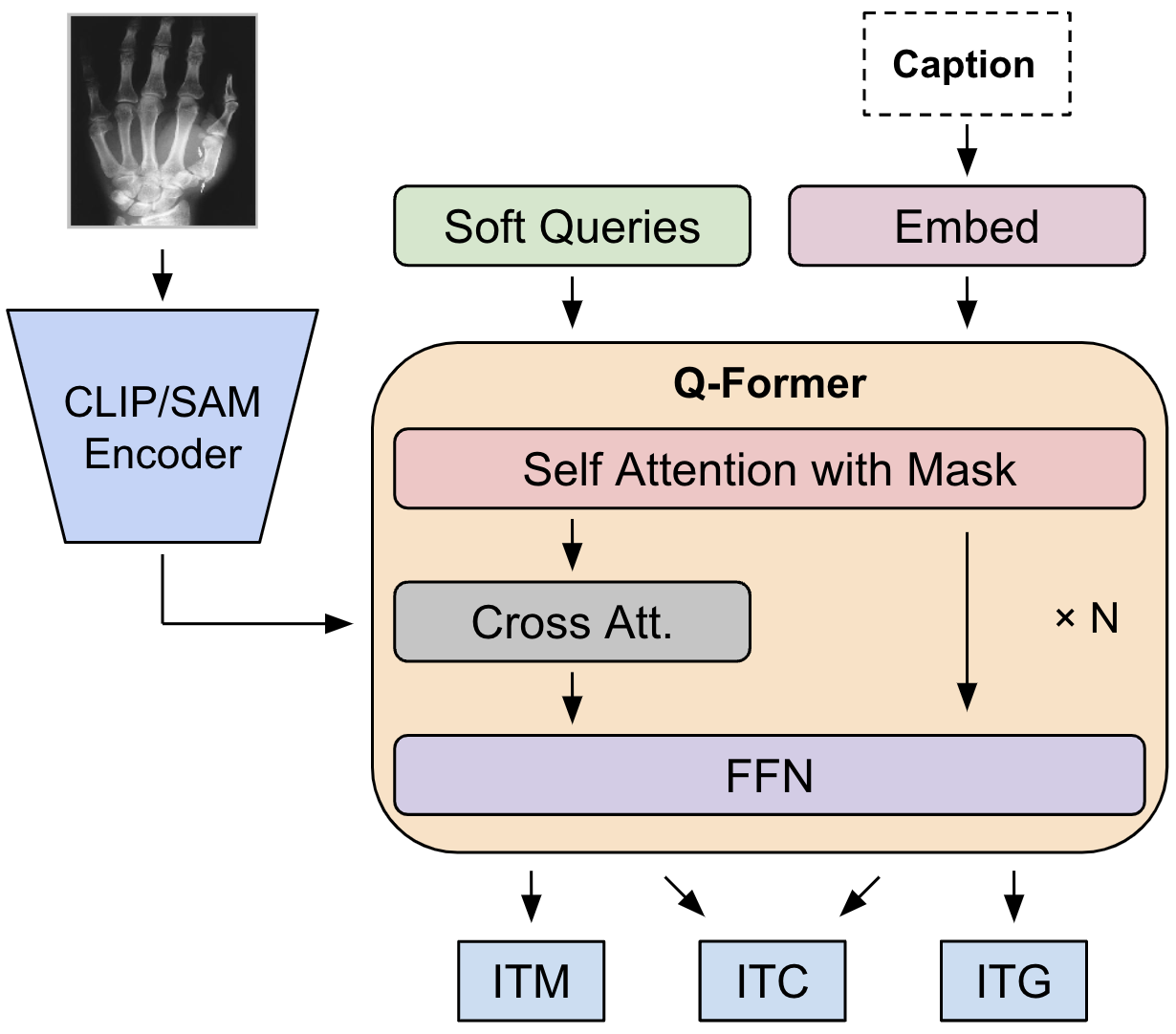}
    \caption{\textbf{Mixed Semantic Pre-training.} For the CLIP Q-Former, we train it using the general datasets. As for the SAM Q-Former, we simultaneously train it on both the general and medical datasets.}
    \vspace{-12pt}
    \label{fig:pre-train}
    
\end{figure}

Models pre-trained using different methods yield distinct granularity and semantic information when extracting features from images. By harnessing the strengths of various pre-training approaches, we adopt a distinctive training strategy to combine general image information and medical domain-specific image information. 
This pre-training strategy is shown in \textbf{Fig. \ref{fig:pre-train}}.

%

In this stage, we separately trained Q-Formers for CLIP and SAM. Initially, we froze the image encoders and fed the images through the encoder to obtain extracted image features. We input both the trainable Soft Queries and embedded image captions into the Q-Former, connecting the image features extracted by the image encoder through Cross Attention to the Q-Former. It's important to note that Cross Attention only operates on the portion corresponding to the Soft Query, and the caption's portion is not used. Taking inspiration from BLIP2\cite{li2023blip2}, we optimized the Q-Former with the following three objectives and used corresponding masks at the Self Attention to meet the different requirements of these three objectives.
\begin{itemize}

\setlength{\itemsep}{1pt}
    \vspace{-6pt}
    \item Image-Text Matching (ITM): Classifying pairs of (image, text) inputs to determine their relevance.
    \vspace{-2pt}
    \item Image-Based Text Generation (ITG): Generating corresponding textual descriptions given image inputs.
    \vspace{-2pt}
    \item Image-Text Contrastive Learning (ITC): Reducing the distance between image and corresponding text features while increasing the distance from irrelevant text features.
    \vspace{-6pt}
\end{itemize}

Due to the need for combining both general semantic information and fine-grained image details in medical image analysis, our model training process involves achieving mixed-semantic representation learning for CLIP and SAM. Given that CLIP excels at capturing more general semantic information, we aim to preserve this capability in our model. Therefore, we load the pre-trained parameters of CLIP Q-Former from BLIP2 \cite{li2023blip2}, which is trained on general image datasets. In contrast, the SAM image encoder has already undergone MAE \cite{he2022masked} pre-training and trained on segmentation tasks\cite{kirillov2023segment}, making it more adept at capturing fine-grained image details compared to CLIP. 
To capture the medical image details, such as pixel-wise semantics, 
we train the Q-Former of SAM using a combination of both general and medical datasets (COCO \cite{lin2015microsoft}, ROCO \cite{pelka2018radiology}, and MedICaT \cite{subramanian2020medicat}). We demonstrate in our subsequent experiments that this training strategy is relatively more effective than other methods, maximizing feature diversity.
\vspace{-8 pt}
\subsection{Captioning with the Frozen LLM}
Due to the mixed semantic pre-training accomplished in the previous stage, the model has already acquired the ability to capture medical image information at both global and local levels and align effectively with text. 
In this stage, we fine-tune the complete model on ROCO and MedICaT datasets with the frozen LLM to generate medical image captions. 
We utilize the OPT \cite{zhang2022opt} as our LLM. Initially, we freeze all parameters of the two image encoders and the LLM, focusing solely on training the Q-Formers and the linear projection layer. Features extracted through the CLIP and SAM encoders, as well as their respective Q-Formers and linear projection layers, are concatenated and fed into the LLM. The model is trained using the LLM loss.

\section{Experiments and Results}
\label{sec:experiments}


\begin{table*}[!t]
    \centering
    \scalebox{0.90}{
        \begin{tabular}{c|ccccccccc}
            \hline
            Models & Bleu 1 & Bleu 2 & Bleu 3 & METEOR & ROUGE\_L & CIDEr & BERT score & BLEURT \\  
            & ($\times 10^3$) & ($\times 10^3$) & ($\times 10^3$) & ($\times 10^3$) & ($\times 10^2$) & ($\times 10^3$) & ($\times 10^2$) & ($\times 10^1$) \\
            \hline
            BLIP2 (G) * & 7.4 & 3.0 & 1.2 & 26.1 & 9.7 & 23.7 & 72.3 & -11.5 \\ 
            BLIP2 (G+M) & 53.2 & 20.6 & 7.5 & 28.3 & 11.1 & 16.4 & 71.6 & -11.5 \\
            SAM-BLIP2 (G+M) & 83.4 & 29.4 & 10.0 & 35.5 & 9.3 & 11.1 & 72.7 & -9.3 \\
            \hline
            MSMedCap (G, G) ** & 101.6 & 45.4 & 22.2 & 59.1 & 15.1 & 56.9 & 76.6 & -7.7 \\
            MSMedCap (G+M, G) & 107.2 & 45.8 & 22.3 & 53.0 & 14.3 & 36.6 & 75.9 & -7.8 \\
            MSMedCap (G+M, G+M) & 48.1 & 22.8 & 10.9 & 49.4 & 14.7 & \textbf{61.0} & 76.6 & -9.1 \\
            \hline
            MSMedCap (G, G+M) & \textbf{108.9} & \textbf{48.1} & \textbf{23.1} & \textbf{62.6} & \textbf{15.4} & 57.5 & \textbf{76.8} & \textbf{-7.6} \\
            \hline
        \end{tabular}
    }
    \caption{Comparison with Benchmarks Across Different Evaluation Metrics. (* The content in parentheses represents the training set for the model; ** The content in parentheses represents the training set for the CLIP Q-Former and the SAM Q-Former, respectively; G = pre-trained on the general datasets, M = pre-trained on the medical datasets; the same below)}
    \label{tab:table}  
    \vspace{-0.2cm}
\end{table*}

\subsection{Experimental Setup}
\textbf{Datasets.} Three widely used datasets are adopted in our experiments, \textit{i.e.}, COCO \cite{lin2015microsoft}, ROCO \cite{pelka2018radiology}, and MedICaT \cite{subramanian2020medicat}. We employ these datasets for the purpose of training and testing. Specifically, the COCO dataset consists of common images, which is referred to as general datasets in the following. The ROCO and MedICaT datasets consist of medical images, which is referred to as medical datasets in the following. To be precise, for the COCO dataset \cite{lin2015microsoft}, we select the train2014 version for training but don't use it to test the output, in the consideration of the fact that general datasets don't match our task. For the medical datasets, we choose 100,000 images from MedICaT \cite{subramanian2020medicat} and 60,000 images from ROCO \cite{pelka2018radiology} for training, and 50,000 images from MedICaT \cite{subramanian2020medicat} and 8,000 images from ROCO \cite{pelka2018radiology} for testing.


\noindent
\textbf{Metrics.}
We employed BLEU \cite{papineni2002bleu}, METEOR \cite{banerjee2005meteor}, ROUGE\_L \cite{lin2004rouge}, CIDEr \cite{vedantam2015cider}, BERTSCORE \cite{zhang2019bertscore}, BARTSCORE \cite{yuan2021bartscore},  and BLEURT\cite{sellam2020bleurt} as the evaluation metrics. For the purpose of convenient comparison, we scale the scores for each metric, which is shown in \textbf{Table \ref{tab:table}}. Higher scores of the metrics indicate more outstanding qualities of generated results.


\noindent
\textbf{Baseline Methods.}\ \ \ 
The optimal model obtained through our approach is referred to as MSMedCap, which is compared against the state-of-the-art model BLIP2 and the revised model SAM-BLIP2, whose CLIP encoder is replaced by SAM encoder. 

\noindent
\textbf{Implementation Details.}\ \ \ 
We conducted three epochs of training for each of the two stages, the mixed semantic pre-training stage and the captioning stage. Furthermore, for our model during the generation process in the finetuning stage, we conducted hyperparameter tuning to guide it and ensure that the generated results are as coherent as possible. 

\subsection{Experimental Design and Results}

To verify the advantage of our architecture and the validity of the training strategy, we perform two sets of experiments. Regarding the naming of the experimental models, it is described in \textbf{Table~\ref{tab:table}}.

\noindent
\textbf{Effectiveness of Model Architecture.} 
We evaluate the output of MSMedCap(G, G+M) and compared it with three models, including BLIP2(G), BLIP2(G+M) and SAM-BLIP2(G+\\M). Our model shows significant improvement compared to the other three models. It seems that the cooperation of the CLIP encoder and SAM encoder is performs better than either of them alone. The CLIP encoder helps capture general semantic information and the SAM encoder helps capture detailed semantic information. Their cooperation significantly improves the quality of the generated text. 

\noindent
\textbf{Effectiveness of Mixed Semantic Pre-training.} 
In the mixed semantic pre-training stage, we design different training strategies for the  CLIP Q-Former and the SAM Q-Former. The ablation study shown in \textbf{Table \ref{tab:table}} verifies the advantage of our training strategy compared to other strategies including MSMedCap(G, G), MSMedCap(G, G+M), MSMedCap(G+M, G+M). The superiority of our model MSMedCap(G, G+M) compared to MSMedCap(G+M, G+M) underscores the effectiveness of mixed semantic training. The superiority of our model MSMedCap(G, G+M) compared to MSMedCap(G+M, G) underscores that the SAM encoder is more proficient at capturing fine-grained medical image details compared to the CLIP encoder.

\begin{figure}[!t]
    \centering
    \includegraphics[width=0.47\textwidth]{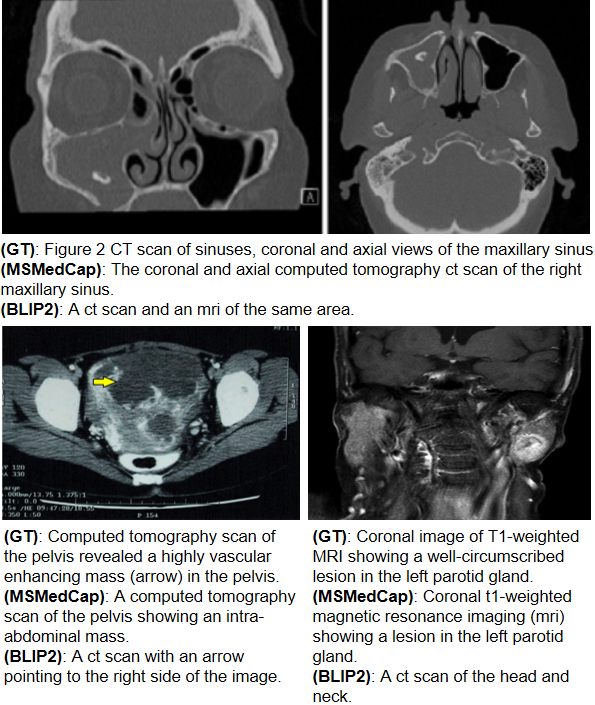}
    \caption{\textbf{Results Visualization.} In our experimental results, we selected several sets of images for visual presentation, comparing our model (MSMedCap) with the baseline (BLIP2 \cite{li2023blip2}) and the ground truth (GT).}
    \label{fig:visual}
    \vspace{-0.4cm}
\end{figure}

\noindent
\textbf{Qualitative Results.}\ \ \ As shown in \textbf{Fig. \ref{fig:visual}}, we selected several representative sets of samples from the testing dataset to demonstrate the effectiveness of our model.

\section{Conclusion}
\label{sec:conclusion}

Facing suboptimal results with BLIP2 in generating pathological diagnoses on medical datasets, we introduced enhancements using the SAM framework. Leveraging the CLIP2 encoder for capturing broader information and the SAM encoder for finer details, we combined both alignment training and text generation on medical datasets. This approach outperformed the original model across various metrics, significantly enhancing output quality. This successful integration addresses the limitations of BLIP2 in providing professional and detailed diagnoses in the medical domain.

Irrelevant details in training captions confuse the model due to matching letter-based labels. We aim to improve accuracy by using a larger language model to refine captions and incorporate medical knowledge. Meanwhile, existing evaluation metrics are inadequate for medical diagnostics as they lack a specialized understanding of medical knowledge. We're considering training-specific evaluation metrics for accurate assessment.





\clearpage
\bibliographystyle{IEEEbib}
\bibliography{refs}

\end{document}